\documentclass[10pt,twocolumn,letterpaper]{article}

\usepackage{iccv}
\usepackage{times}
\usepackage{epsfig}
\usepackage{graphicx}
\usepackage{subfigure}
\usepackage{amsmath}
\usepackage{amssymb}
\usepackage{booktabs, multirow} 
\usepackage{soul}
\usepackage[table]{xcolor} 
\usepackage{changepage,threeparttable} 
\usepackage[numbers,sort,compress]{natbib}
\newcommand{\best}[1]{{\color{blue}{\textbf{#1}}}} 
\newcommand{\second}[1]{{\color{black}{\textbf{#1}}}}
\newcommand{\ours}{{MUSIQ}}
\newcommand{\fullours}{{MUSIQ-single}}

\usepackage[pagebackref=true,breaklinks=true,colorlinks,citecolor=blue,linkcolor=blue,bookmarks=false]{hyperref}

\iccvfinalcopy 

\def\iccvPaperID{10958} 

\ificcvfinal\fi

\begin{document}

\title{MUSIQ: Multi-scale Image Quality Transformer}

\author{
Junjie Ke$^{1}$, Qifei Wang$^{1}$, Yilin Wang$^{2}$, Peyman Milanfar$^{1}$, Feng Yang$^{1}$ \\
$^{1}$Google Research, $^{2}$Google\\
{\tt\small  {\{junjiek, qfwang, yilin, milanfar, fengyang\}@google.com}}
}

\maketitle
\ificcvfinal\thispagestyle{empty}\fi

\begin{abstract}
Image quality assessment (IQA) is an important research topic for understanding and improving visual experience. The current state-of-the-art IQA methods are based on convolutional neural networks (CNNs). The performance of CNN-based models is often compromised by the fixed shape constraint in batch training. To accommodate this, the input images are usually resized and cropped to a fixed shape, causing image quality degradation. To address this, we design a multi-scale image quality Transformer (MUSIQ) to process native resolution images with varying sizes and aspect ratios. With a multi-scale image representation, our proposed method can capture image quality at different granularities. Furthermore, a novel hash-based 2D spatial embedding and a scale embedding is proposed to support the positional embedding in the multi-scale representation. Experimental results verify that our method can achieve state-of-the-art performance on multiple large scale IQA datasets such as PaQ-2-PiQ~\cite{ying2020patches},  SPAQ~\cite{fang2020perceptual}, and KonIQ-10k~\cite{hosu2020koniq}. \footnote{Checkpoints and code are available at \url{https://github.com/google-research/google-research/tree/master/musiq}}
\end{abstract}

\section{Introduction}
\begin{figure}[tp!]
\centering
\includegraphics[width=8cm]{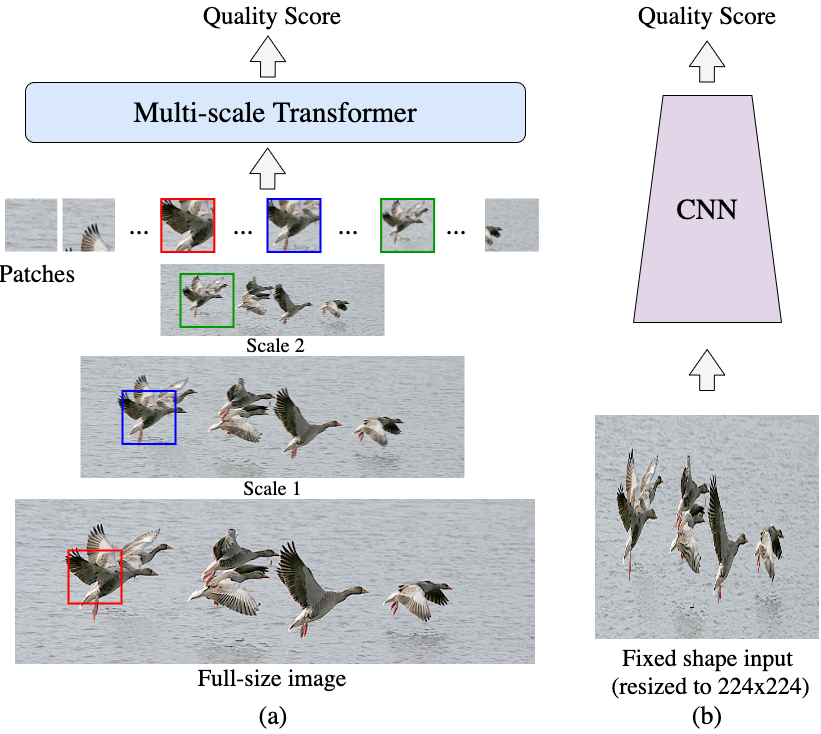}\vspace{-0.5mm}
\caption{In CNN-based models (b), images need to be resized or cropped to a fixed shape for batch training. However, such preprocessing can alter image aspect ratio and composition, thus impacting image quality. Our patch-based \ours\ model (a) can process the full-size image and extract multi-scale features, which aligns with the human visual system.}\vspace{-2.5mm}
\label{fig:teaser} 
\end{figure}

\noindent The goal of image quality assessment (IQA) is to quantify perceptual quality of images. In the deep learning era, many IQA approaches \cite{fang2020perceptual, su2020blindly, talebi2018nima, ying2020patches, zhang2018blind} have achieved significant success by leveraging the power of convolutional neural networks (CNNs). However, the CNN-based IQA models are often constrained by the fixed-size input requirement in batch training, ~\ie, the input images need to be resized or cropped to a fixed shape as shown in Figure \ref{fig:teaser} (b).  This preprocessing is problematic for IQA because images in the wild have varying aspect ratios and resolutions. Resizing and cropping can impact image composition or introduce distortions, thus changing the quality of the image.

To learn IQA on the full-size image, the existing CNN-based approaches use either adaptive pooling or resizing to get a fixed-size convolutional feature map. MNA-CNN \cite{mai2016composition} processes a single image in each training batch which is not practical for training on a large dataset. \citet{hosu2019effective} extracts and stores fixed-size features offline, which costs additional storage for every augmented image. To keep aspect ratio, \citet{chen2020adaptive} proposes a dedicated convolution to preserve aspect ratio in the convolutional receptive field. Its evaluation verifies the importance of aspect-ratio-preserving (ARP) in the IQA tasks. But it still needs resizing and smart grouping for effective batch training. 

In this paper, we propose a patch-based multi-scale image quality Transformer (\ours) to bypass the CNN constraints on fixed input size and predict the quality effectively on the native resolution image as shown in Figure~\ref{fig:teaser} (a). Transformer~\cite{NIPS2017_3f5ee243} is first proposed for natural language processing (NLP) and has recently been studied for various vision tasks~\cite{carion2020end, chen2020pre, pmlr-v119-chen20s, dosovitskiy2020}. Among these, the Vision Transformer (ViT) \cite{dosovitskiy2020} splits each image into a sequence of fixed-size patches, encodes each patch as a token, and then applies Transformer to the sequence for image classification. In theory, such kind of patch-based Transformer models can handle arbitrary numbers of patches (up to memory constraints), and therefore do not require preprocessing the input image to a fixed resolution. This motivates us to apply the patch-based Transformer on the IQA tasks with the full-size images as input.

Another aspect for improving IQA models is to imitate the human visual system which captures an image in a multi-scale fashion~\cite{adelson1984pyramid}. Previous works  \cite{hosu2019effective, lin2017feature, zhang2018unreasonable} have shown the benefit of using multi-scale features extracted from CNN feature maps at different depths. This inspires us to transform the native resolution image into a multi-scale representation, enabling the Transformer's self-attention mechanism to capture information on both fine-grained detailed patches and coarse-grained global patches. Besides, unlike the convolution operation in CNNs that has a relatively limited receptive field, self-attention can attend to the whole input sequence and it can therefore effectively capture the image quality at different granularities. 

However, it is not straightforward to apply the Transformer on the multi-aspect-ratio multi-scale input. Although self-attention accepts arbitrary length of the input sequence, it is permutation-invariant and therefore cannot capture patch location in the image. To mitigate this, ViT~\cite{dosovitskiy2020} adds fixed-length positional embedding to encode the absolute position of each patch in the image. However, the fixed-length positional encoding fails when the input length varies. To solve this issue, we propose a novel hash-based 2D spatial embedding that maps the patch positions to a fixed grid to effectively handle images with arbitrary aspect ratios and resolutions. Moreover, since the patch locations at each scale are hashed to the same grid, it aligns spatially close patches at different scales so that the Transformer model can leverage information across multiple scales. In addition to the spatial embedding, a separate scale embedding is further introduced to help the Transformer distinguish patches coming from different scales in the multi-scale representation.

The main contributions of this paper can be summarized into three-folds:
\begin{itemize}
\vspace{-1mm}
\setlength\itemsep{-1mm}
    \item  We propose a patch-based multi-scale image quality Transformer (MUSIQ), which supports processing full-size input with varying aspect ratios or resolutions, and allows multi-scale feature extraction.
    
    \item A novel hash-based 2D spatial embedding and a scale embedding are proposed to support positional encoding in the multi-scale representation, helping the Transformer capture information across space and scales.

    \item We apply \ours\ on four large-scale IQA datasets. It consistently achieves the state-of-the-art performance on three technical quality datasets: PaQ-2-PiQ~\cite{ying2020patches}, KonIQ-10k~\cite{hosu2020koniq}, and SPAQ \cite{fang2020perceptual},  and is on-par with the state-of-the-art on the aesthetic quality dataset AVA~\cite{murray2012ava}.
\vspace{-1mm}
\end{itemize}

\begin{figure*}[tp!]
\centering
\includegraphics[width=17.3cm]{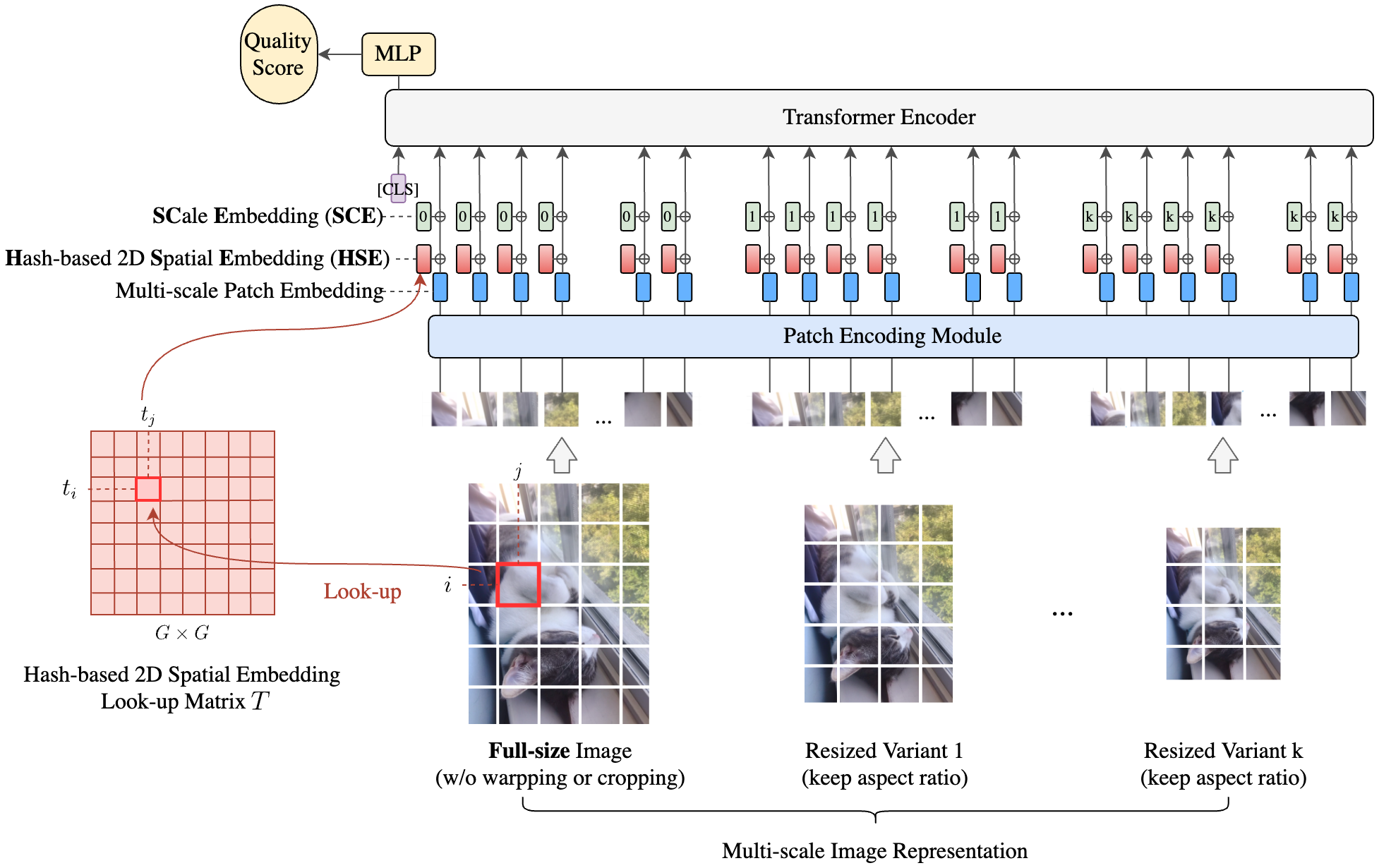}\vspace{-0.5mm}
\caption{Model overview of \ours. We construct a multi-scale image representation as input, including the native resolution image and its ARP resized variants. Each image is split into fixed-size patches which are embedded by a patch encoding module (blue boxes). To capture 2D structure of the image and handle images of varying aspect ratios, the spatial embedding is encoded by hashing the patch position $(i,j)$ to $(t_i, t_j)$ within a grid of learnable embeddings (red boxes). Scale Embedding (green boxes) is introduced to capture scale information. The Transformer encoder takes the input tokens and performs multi-head self-attention. To predict the image quality, we follow a common strategy in Transformers to add an [CLS] token to the sequence to represent the whole multi-scale input and use the corresponding Transformer output as the final representation. }\vspace{-2.5mm}
\label{fig:overview} 
\end{figure*}

\section{Related Work}
\noindent\textbf{Image Quality Assessment.} Image quality assessment aims to quantitatively predict perceptual image quality. There are two important aspects for assessing image quality: technical quality \cite{hosu2020koniq} and aesthetic quality \cite{murray2012ava}. The former focuses on perceptual distortions while the latter also relates to image composition, artistic value and so on. In the past years, researchers proposed many IQA methods: early natural scene statistics based~\cite{ghadiyaram2017perceptual, mittal2012no, moorthy2011blind, zhang2015feature}, codebook-based~\cite{xue2013learning, ye2012unsupervised} and CNN-based~\cite{fang2020perceptual, su2020blindly, talebi2018nima, ying2020patches, zhang2018blind}. CNN-based methods achieve the state-of-the-art performance. However they usually need to crop or resize images to a fixed size in batch training, which affects the image quality. Several methods have been proposed to mitigate the distortion from resizing and cropping in CNN-based IQA. An ensemble of multi-crops from the original image is proven to be effective for IQA \cite{chen2020adaptive, hosu2019effective, ma2017lamp, sheng2018attention, su2020blindly}, but it introduces non-negligible inference cost. In addition, MNA-CNN \cite{mai2016composition} handles full-size input by adaptively pooling the feature map to a fixed shape. However, it only accepts a single input image for each training batch to preserve the original resolution which is not efficient for large scale training. \citet{hosu2019effective} extracted and stored the fixed-sized features from the full-size image for model training which costs extra storage for every augmented image and is inefficient for large scale training. \citet{chen2020adaptive} proposed an adaptive fractional dilated convolution to adapt the receptive field according to the image aspect ratio. The method preserves aspect ratio but cannot handle full-size input without resizing. It also needs smart grouping strategy in mini-batch training.

\noindent\textbf{Transformers in Vision.} Transformers \cite{NIPS2017_3f5ee243} were first applied to NLP tasks and achieved great performance \cite{yang2019xlNet, devlin2018bert, liu2019roberta}. Recent works applied transformers on various vision tasks \cite{carion2020end, chen2020pre, pmlr-v119-chen20s, dosovitskiy2020}. Among these, the Vision Transformer (ViT) \cite{dosovitskiy2020} employs a pure Transformer architecture to classify images by treating an image as a sequence of patches. 
For batch training, ViT resizes the input images to a fixed squared size,~\eg, 224 $\times$ 224, where fixed number of patches are extracted and combined with fixed-length positional embedding. This constrains its usage for IQA since resizing will affect the image quality. To solve this, we propose a novel Transformer-based architecture that accepts the full-size image for IQA.

\noindent\textbf{Positional Embeddings.} Positional embeddings are introduced in Transformers to encode the order of the input sequence \cite{NIPS2017_3f5ee243}. Without it, the self-attention operation is permutation-invariant \cite{bello2019attention}. \citet{NIPS2017_3f5ee243} used deterministic positional embeddings generated from sinusoidal functions. ViT \cite{dosovitskiy2020} showed that the deterministic and learnable positional embeddings \cite{gehring2017convolutional} works equally well. However, those positional embeddings are generated for fixed-length sequences. When the input resolution changes, the pre-trained positional embeddings is no longer meaningful. Relative positional embeddings~\cite{shaw2018self, bello2019attention} is proposed to encode relative distance instead of absolute position. Although the relative positional embeddings can work for variable length inputs, it requires substantial modifications in Transformer attention and cannot capture multi-scale positions in our use case.

\section{Multi-scale Image Quality Transformer}

\subsection{Overall Architecture}
\noindent To tackle the challenge of learning IQA on full-size images, we propose a multi-scale image quality Transformer (\ours) which can handle inputs with arbitrary aspect ratios and resolutions. An overview of the model is shown in Figure~\ref{fig:overview}.

We first make a multi-scale representation of the input image, containing the native resolution image and its ARP resized variants. The images at different scales are partitioned into fixed-size patches and fed into the model. Since patches are from images of varying resolutions, we need to effectively encode the multi-aspect-ratio multi-scale input into a sequence of tokens (the small boxes in Figure \ref{fig:overview}), capturing both the pixel, spatial, and scale information.

To achieve this, we design three encoding components in \ours, including: 1) A patch encoding module to encode patches extracted from the multi-scale representation (Section \ref{sec:multi-scale-input}); 2) A novel hash-based spatial embedding module to encode the 2D spatial position for each patch (Section \ref{sec:hse}); 3) A learnable scale embedding to encode different scale (Section \ref{sec:sce}). 

After encoding the multi-scale input into a sequence of tokens, we use the standard approach of prepending an extra learnable ``classification token" (CLS) \cite{devlin2018bert, dosovitskiy2020}. The CLS token state at the output of the Transformer encoder serves as the final image representation. We then add a fully connected layer on top to predict the image quality score. Since \ours\ only changes the input encoding, it is compatible with any Transformer variants. To demonstrate the effectiveness of the proposed method, we use the classic Transformer \cite{NIPS2017_3f5ee243} (Appendix A) with a relatively lightweight setting to make model size comparable to ResNet-50 in our experiments.

\subsection{Multi-scale Patch Embedding}
\label{sec:multi-scale-input}
\noindent Image quality is affected by both the local details and global composition. In order to capture both the global and local information, we propose to model the input image with a multi-scale representation. Patches from different scales enables the Transformer to aggregate information across multiple scales and spatial locations. 

As shown in Figure \ref{fig:overview}, the multi-scale input is composed of the full-size image with height $H$, width $W$, channel $C$, and a sequence of ARP resized images from the full-size image using Gaussian kernel. The resized images have height $ h_k$, width $w_k$, channel $C$, where $k=1,..., K$ and $K$ is the number of resized variants for each input. To align resized images for a consistent global view, we fix the longer side length to $L_k$ for each resized variant and yield:
\vspace{-1mm}
\begin{equation}
\alpha_k = L_k/\max(H, W),\ \ h_k = \alpha_kH,\ \ w_k = \alpha_kW
\end{equation}
$\alpha_k$ represents the resizing factor for each scale.

Square patches with size $P$ are extracted from each image in the multi-scale representation. For images whose width or height are not multiples of $P$, we pad the image with zeros accordingly. Each patch is encoded into a $D$-dimension embedding by the patch encoder module. $D$ is the latent token size used in the Transformer.

Instead of encoding the patches with a linear projection as in~\cite{dosovitskiy2020}, we choose a 5-layer ResNet~\cite{he2016deep} with a fully connected layer of size $D$ as the patch encoder module to learn a better representation for the input patch. We find that encoding the patch with a few convolution layers performs better than linear projection when pre-training on ILSVRC-2012 ImageNet \cite{imagenet} (see Section \ref{sec:ablation-study}). Since the patch encoding module is lightweight and shared across all the input patches whose size $P$ is small, it only adds a small amount of parameters.

The sequence of patch embeddings output from the patch encoder module are concatenated together to form a multi-scale embedding sequence for the input image. The number of patches from the original image and the resized ones are calculated as $N={HW}/{P^2}$ and $n_k ={h_kw_k}/{P^2}$.

Since each input image has a different resolution and aspect ratio, $H$ and $W$ are different for each input and therefore $N$ and $n_k$ are different. To get fixed-length input during training, we follow the common practice in NLP \cite{NIPS2017_3f5ee243} to zero-pad the encoded patch tokens to the same length. An input mask is attached to indicate the effective input, which will be used in the Transformer to perform masked self-attention (Appendix A.3). Note that the padding operation will not change the input because the padding tokens are ignored in the multi-head attention by masking them.

As previously mentioned, we fix the longer length to $L_k$ for each resized variant. Therefore $n_k \leq L_k^2/{P^2} = m_k$ and we can safely pad to $m_k$. For the native resolution image, we simply pad or cut the sequence to a fixed length $l$. The padding is not necessary during single-input evaluation because the sequence length can be arbitrary.

\subsection{Hash-based 2D Spatial Embedding}
\label{sec:hse}
\noindent Spatial positional embedding is important in vision Transformers to inject awareness of the 2D image structure in the 1D sequence input \cite{dosovitskiy2020}. The traditional fixed-length positional embedding assigns an embedding for every input location. This fails for variable input resolutions where the number of patches are different and therefore each patch in the sequence may come from an arbitrary location in the image. Besides, the traditional positional embedding models each position independently and therefore it cannot align the spatially close patches from different scales.

We argue that an effective spatial embedding design for \ours\ should meet the following requirements: 1) effectively encode patch spatial information under different aspect ratios and input resolutions; 2) spatially close patches at different scales should have close spatial embeddings; 3) efficient and easy to implement, non-intrusive to the Transformer attention.

Based on that, we propose a novel hash-based 2D spatial embedding (HSE) where the patch locating at row $i$, column $j$ is hashed to the corresponding element in a $G\times{G}$ grid. Each element in the grid is a $D$-dimensional embedding.

We define HSE by a learnable matrix  $T \in \mathbb{R}^{G\times G\times D}$. Suppose the input resolution is $H \times W$. The input image will be partitioned into $\frac{H}{P} \times \frac{W}{P}$ patches. For the patch at position $(i, j)$, its spatial embedding is defined by the element at position $(t_i, t_j)$ in $T$ where
\vspace{-2mm}
\begin{equation}
    t_i = \frac{i\times G}{H/P},\ t_j = \frac{j\times G}{W/P}
\end{equation}

The $D$-dimensional spatial embedding $T_{t_i, t_j}$ is added to the patch embedding element-wisely as shown in Figure~\ref{fig:overview}. For fast lookup, we simply round $(t_i, t_j)$ to the nearest integers. HSE does not require any changes in the Transformer attention module. Moreover, both the computation of $t_i$ and $t_j$ and the lookup are lightweight and easy to implement. 

To align patches across scales, patch locations from all scales are mapped to the same grid $T$. As a result, patches located closely in the image but from different scales are mapped to spatially close embeddings in $T$, since $i$ and $H$ as well as $j$ and $W$ change proportionally to the resizing factor $\alpha$. This achieves spatial alignment across different images from the multi-scale representation.

There is a trade-off between expressiveness and train-ability with the choice hash grid size $G$. Small $G$ may result in a lot of collision between patches which makes the model unable to distinguish spatially close patches. Large $G$ wastes memory and may need more diverse resolutions to train. In our IQA setting where rough positional information is sufficient, we find once $G$ is large enough, changing $G$ only results in small performance differences (see Appendix B). We set $G=10$ in the experiments.

\subsection{Scale Embedding}
\label{sec:sce}
\noindent Since we reuse the same hashing matrix for all images, HSE does not make a distinction between patches from different scales. Therefore, we introduce an additional scale embedding (SCE) to help the model effectively distinguish information coming from different scales and better utilize information across scales. In other words, SCE marks which input scale the patch is coming from in the multi-scale representation.

We define SCE as a learnable scale embedding $Q \in \mathbb{R}^{(K+1) \times D}$ for the input image with $K$-scale resized variants. Following the spatial embedding, the first element $Q_0 \in \mathbb{R}^D$ is added element-wisely to all the $D$-dimensional patch embeddings from the native resolution image. $Q_k \in \mathbb{R}^D, k=1,...,K$ are also added element-wisely to all the  patch embeddings from the resized image at scale $k$.

\subsection{Pre-training and Fine-tuning}

\noindent Typically, the Transformer models need to be pre-trained on the large datasets, \eg~ImageNet, and fine-tuned on the downstream tasks. During the pre-training, we still keep random cropping as an augmentation to generate images of different sizes. However, instead of doing square resizing like the common practice in image classification, we intentionally skip resizing to prime the model for inputs with different resolutions and aspect ratios. We also employ common augmentations such as RandAugment \cite{cubuk2020randaugment} and mixup \cite{zhang2017mixup} in pre-training.

When fine-tuning on IQA tasks, we do not resize or crop the input image to preserve the image composition and aspect ratio. In fact, we only use random horizontal flipping for augmentation in fine-tuning. For evaluation, our method can be directly applied on the original image without aggregating multiple augmentations (\eg multi-crops sampling).

When fine-tuning on the IQA datasets, we use common regression losses such as L1 loss for single mean opinion score (MOS) and Earth Mover Distance (EMD) loss to predict the quality score distribution \cite{talebi2018nima}:
\vspace{-4mm}
\begin{equation}
    EMD (p, \hat{p}) = (\frac{1}{N} \sum_{m=1}^N | CDF_p(m) - CDF_{\hat{p}}(m)|^r)^{\frac{1}{r}}
\vspace{-3mm}
\end{equation}
where $p$ is the normalized score distribution and $CDF_p(m)$ is the cumulative distribution function as $\sum_{i=1}^m p_i$.

\section{Experimental Results}

\subsection{Datasets}
\label{sec:datasets}
\noindent We run experiments on four large-scale image quality datasets including three technical quality datasets (PaQ-2-PiQ \cite{ying2020patches}, SPAQ \cite{fang2020perceptual},  KonIQ-10k \cite{hosu2020koniq}) and one aesthetics quality dataset (AVA~\cite{murray2012ava}).

PaQ-2-PiQ is the largest picture technical quality dataset by far which contains 40k real-world images and 120k cropped patches. Each image or patch is associated with a MOS. Since our model does not make a distinction between image and extracted patches, we simply use all the 30k full-size images and the corresponding 90k patches from the training split to train the model. We then run the evaluation on the 7.7k full-size validation and 1.8k test set.

SPAQ dataset consists of 11k pictures taken by 66 smartphones. For a fair comparison, we follow \cite{fang2020perceptual} to resize the raw images such that the shorter side is 512. We only use the image and its corresponding MOS for training, not including the extra tag information in the dataset.

KonIQ-10k contains 10k images selected from a large public multimedia database YFCC100M~\cite{thomee2016yfcc100m}.

AVA is an image aesthetic assessment dataset. It contains 250k images with 10-scale score distribution for each.

For KonIQ-10k, we follow \cite{zhu2020metaiqa, su2020blindly} to randomly sample 80\% images for each run and report the results on the remaining 20\%. For other datasets, we use the same split as the previous literature.

\subsection{Implementation Details}
\noindent For \ours, the multi-scale representation is constructed as the native resolution image and two ARP resized input ($L_1=224$ and $L_2=384$) by default. It therefore uses 3-scale input. Our method also works on 1-scale input using just the full-size image without resized variants. We report the results of this single-scale setting as \fullours.

We use patch size $P = 32$. The dimensions for Transformer input tokens are $D=384$, which is also the dimension for pixel patch embedding, HSE and SCE. The grid size of HSE is set to $G=10$. We use the classic Transformer \cite{NIPS2017_3f5ee243} with lightweight parameters (384 hidden size, 14 layers, 1152 MLP size and 6 heads) to make the model size comparable to ResNet-50. The final model has around 27 million total parameters.

We pre-train our models on ImageNet for 300 epochs, using Adam with $\beta_1=0.9, \beta_2 =0.999$, a batch size of $4096$, $0.1$ weight decay and cosine learning rate decay from $0.001$. We set the maximum number of patches from full-size image $l$ to 512 in training. For fine-tuning, we use SGD with momentum and cosine learning rate decay from $0.0002, 0.0001, 0.0001, 0.12$ for 10, 30, 30, 20 epochs on PaQ-2-PiQ, KonIQ-10k, SPAQ, and AVA, respectively. Batch size is set to 512 for AVA, 96 for KonIQ-10k, and 128 for the rest. For AVA, we use the EMD loss with $r=2$. For other datasets, we use the $L1$ loss.

The models are trained on TPUv3. All the results from our method are averaged across 10 runs. Spearman rank ordered correlation (SRCC), Pearson linear correlation (PLCC), and the standard deviation (std) are reported.

\subsection{Comparing with the State-of-the-art (SOTA)}
\noindent\textbf{Results on PaQ-2-PiQ.} Table \ref{tab:paq2piq-results} shows the results on the PaQ-2-PiQ dataset. Our proposed \ours\ outperforms other methods on both the validation and test sets. Notably, the test set is entirely composed of pictures having at least one dimension exceeding 640 \cite{ying2020patches}. This is very challenging for traditional deep learning approaches where resizing is inevitable. Our method is able to outperform previous methods by a large margin on the full-size test set which verifies its robustness and effectiveness. 

\begin{table}[!tp]
\begin{center}
\footnotesize
\begin{tabular}{lccccc}\toprule
&\multicolumn{2}{c}{Validation Set} &\multicolumn{2}{c}{Test Set} \\\cmidrule{2-5}
method &SRCC &PLCC &SRCC &PLCC \\\midrule
BRISQUE \cite{mittal2012no} &0.303 &0.341 &0.288 &0.373 \\
NIQE \cite{mittal2012making} &0.094 &0.131 &0.211 &0.288 \\
CNNIQA \cite{kang2014convolutional} &0.259 &0.242 &0.266 &0.223 \\
NIMA \cite{talebi2018nima} &0.521 &0.609 &0.583 &0.639 \\
Ying \etal \cite{ying2020patches} &0.562 &0.649 &0.601 &0.685 \\\midrule
\fullours  &\second{0.563} &\second{0.651} &\second{0.640} &\second{0.721} \\
\ours\ (Ours) &\best{0.566} &\best{0.661} &\best{0.646} &\best{0.739} \\
std &$\pm 0.002$ &$\pm 0.003$ &$\pm 0.005$ &$\pm 0.006$ \\
\bottomrule
\end{tabular}
\end{center}
\vspace{-2mm}
\caption{Results on PaQ-2-PiQ full-size validation and test sets. Blue and black numbers in bold represent the best and second best respectively. We take numbers from \cite{ying2020patches} for the results of the reference methods.} \label{tab:paq2piq-results}
\vspace{-2mm}
\end{table}

\begin{table}[!tp]
\begin{center}
\footnotesize
\begin{tabular}{lccc}\toprule
method &SRCC &PLCC \\\midrule
BRISQUE \cite{mittal2012no} &0.665 &0.681 \\
ILNIQE \cite{zhang2015feature} &0.507 &0.523 \\
HOSA \cite{xu2016blind} &0.671 &0.694 \\
BIECON \cite{kim2016fully} &0.618 &0.651 \\
WaDIQaM \cite{bosse2017deep} &0.797 &0.805 \\
PQR \cite{zeng2017probabilistic} &0.880 &0.884 \\
SFA \cite{li2018has} &0.856 &0.872 \\
DBCNN \cite{zhang2018blind} &0.875 &0.884 \\
MetaIQA \cite{zhu2020metaiqa}  &0.850 &0.887 \\
BIQA \cite{su2020blindly} (\textbf{25} crops) &\second{0.906} &0.917 \\\midrule
\fullours &0.905 &\second{0.919} \\
\ours\ (Ours) &\best{0.916} &\best{0.928} \\
std &$\pm 0.002$ &$\pm 0.003$ \\
\bottomrule
\end{tabular}
\end{center}
\vspace{-2mm}
\caption{Results on KonIQ-10k dataset. Blue and black numbers in bold represent the best and second best respectively. We take numbers from \cite{su2020blindly, zhu2020metaiqa} for results of the reference methods.} \label{tab:koniq-results}
\vspace{-3mm}
\end{table}

\noindent\textbf{Results on KonIQ-10k.} Table \ref{tab:koniq-results} shows the results on the KonIQ-10k dataset. Our method outperforms the SOTA methods. In particular, BIQA \cite{su2020blindly} needs to sample 25 crops from each image during training and testing. This kind of multi-crops ensemble is a way to mitigate the fixed shape constraint in the CNN models. But since each crop is only a sub-view of the whole image, the ensemble is still an approximate approach. Moreover, it adds additional inference cost for every crop and sampling can introduce randomness in the result. Since \ours\ takes the full-size image as input, it can directly learn the best aggregation of information across the full image and only one evaluation is involved.

\noindent\textbf{Results on SPAQ.} Table \ref{tab:spaq-results} shows the results on the SPAQ dataset. Overall, our model is able to outperform other methods in terms of both SRCC and PLCC. 

\begin{table}[!tp]
\begin{center}
\footnotesize
\begin{tabular}{lrrr}\toprule
method &SRCC &PLCC \\\midrule
DIIVINE \cite{moorthy2011blind} &0.599 &0.600 \\
BRISQUE \cite{mittal2012no} &0.809 &0.817 \\
CORNIA \cite{ye2012unsupervised} &0.709 &0.725 \\
QAC \cite{xue2013learning} &0.092 &0.497 \\
ILNIQE \cite{zhang2015feature} &0.713 &0.721 \\
FRIQUEE \cite{ghadiyaram2017perceptual} &0.819 &0.830 \\
DBCNN \cite{zhang2018blind} &0.911 &0.915 \\
Fang \etal \cite{fang2020perceptual} (w/o extra info) &0.908 &0.909 \\\midrule
\fullours &\best{0.917} &\second{0.920} \\
\ours\ (Ours) &\best{0.917} &\best{0.921} \\
std &$\pm 0.002$ &$\pm 0.002$ \\
\bottomrule
\end{tabular}
\end{center}
\vspace{-2mm}
\caption{Results on SPAQ dataset. Blue and black numbers in bold  represent the best and second best respectively. We take numbers from \cite{fang2020perceptual} for results of the reference methods.}\label{tab:spaq-results}
\vspace{-2mm}
\end{table}

\begin{table}[!tp]
\footnotesize
\begin{center}
\setlength\tabcolsep{2.5pt}
\begin{tabular}{lcccc}
\toprule
method & cls. acc. & MSE $\downarrow$ & SRCC & PLCC \\\midrule
MNA-CNN-Scene \cite{mai2016composition} & 0.765 & - & - & - \\
Kong \etal \cite{kong2016photo} & 0.773 & - & 0.558 & - \\
AMP \cite{murray2017deep} & 0.803 & 0.279 & 0.709 & - \\
A-Lamp \cite{ma2017lamp} (\textbf{50} crops) & 0.825 & - & - & - \\
NIMA (VGG16) \cite{talebi2018nima} & 0.806 & - & 0.592 & 0.610 \\
NIMA (Inception-v2) \cite{talebi2018nima} & 0.815 & - & 0.612 & 0.636 \\
$MP_{ada}$ \cite{sheng2018attention} ( $\ge$ \textbf{32} crops) & 0.830 & - & - & - \\
Zeng \etal (ResNet101) \cite{zeng2019unified} & 0.808 & 0.275  & 0.719 & 0.720 \\
Hosu \etal \cite{hosu2019effective} (\textbf{20} crops) & 0.817 & - & \best{0.756} & \best{0.757} \\
AFDC + SPP (single warp) \cite{chen2020adaptive} & \second{0.830} & 0.273 & 0.648 & - \\
AFDC + SPP (\textbf{4} warps) \cite{chen2020adaptive} & \best{0.832} &0.271 & 0.649 & 0.671 \\\midrule
\fullours &0.814 &\second{0.247} &0.719 &0.731 \\
\ours\ (Ours) & 0.815 & \best{0.242} & \second{0.726} & \second{0.738} \\
std &$\pm 0.121$ &$\pm 0.001$ &$\pm 0.001$ &$\pm 0.001$ \\
\bottomrule
\end{tabular}
\end{center}
\vspace{-2mm}
\caption{Results on AVA dataset. Blue and black numbers in bold represent the best and second best respectively. cls. acc. stands for classification accuracy. MSE stands for mean square error. We take numbers from \cite{chen2020adaptive} for results of the reference methods.}\label{tab:ava-results}
\vspace{-3mm}
\end{table}

\noindent\textbf{Results on AVA.} Table \ref{tab:ava-results} shows the results on the AVA dataset. Our method achieves the best MSE and has top SRCC and PLCC. As previously discussed, instead of multi-crops sampling, our model can accurately predict image aesthetics by directly looking at the full-size image.

\subsection{Ablation Studies}\label{sec:ablation-study}

\noindent\textbf{Importance of Aspect-Ratio-Preserving (ARP).} CNN-based IQA models usually resize the input image to a square resolution without preserving the original aspect ratio. We argue that such preprocessing can be detrimental to IQA because it alters the image composition. To verify that, we compare the performance of the proposed model with either square or ARP resizing. As shown in Table \ref{tab:arp-resizing}, ARP resizing performs better than square resizing, demonstrating the importance of ARP when assessing image quality.

\begin{table}[!tp]
\footnotesize
\begin{center}
\setlength\tabcolsep{1.0pt}
\begin{tabular}{lcccc}\toprule
method & \# Params &SRCC &PLCC \\\midrule
NIMA(Inception-v2) \cite{talebi2018nima} (224 square input) &56M &0.612 &0.636 \\
NIMA(ResNet50)* (384 square input) &24M &0.624 &0.632 \\\midrule
ViT-Base 32* (384 square input) \cite{dosovitskiy2020} &88M &0.654 &0.664 \\
ViT-Small 32* (384 square input) \cite{dosovitskiy2020} &22M &0.656 &0.665 \\\midrule
\ours\ w/ square resizing (512, 384, 224) &27M &0.706 &0.720 \\
\ours\ w/ ARP resizing (512, 384, 224) &27M &0.712 &0.726 \\
\ours\ w/ ARP resizing (full, 384, 224) &27M &\textbf{0.726} &\textbf{0.738} \\
\bottomrule
\end{tabular}
\end{center}
\vspace{-1mm}
\caption{Comparison of ARP resizing and square resizing on AVA dataset. * means our implementation. ViT-Small* is constructed by replacing the Transformer backbone in ViT with our 384-dim lightweight Transformer. The last group of rows show our method with different resizing methods. Numbers in the bracket show the resolution used in the multi-scale representation. }\label{tab:arp-resizing}\vspace{-3mm}
\end{table}

\begin{figure}[!tp]
\centering
\includegraphics[width=8.5cm]{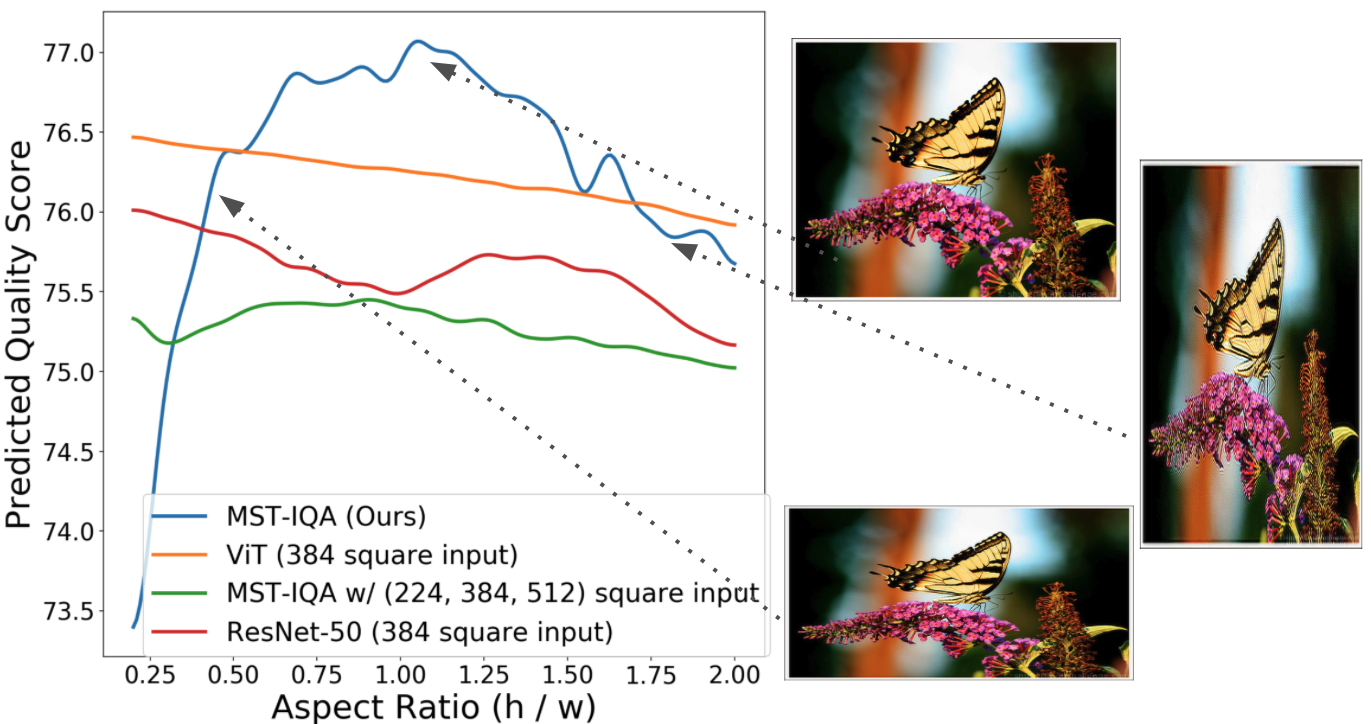}
\caption{Model predictions for an image resized to different aspect ratios. The blue curve shows \ours\ with ARP resizing. The green curve shows our model trained and evaluated with square input. Orange and red curves show the ViT and ResNet-50 with square input. \ours\ can detect quality degradation due to unnatural resizing while other methods are not sensitive.  }\vspace{-1mm}
\label{fig:butterfly_aspect_ratio} 
\end{figure} 

To intuitively understand the importance of keeping aspect ratios in IQA, we follow \cite{chen2020adaptive} to artificially resize the same image into different aspect ratios and run models to predict quality scores. Since aggressive resizing will cause image quality degradation, a good IQA model should give lower scores to such unnatural looking images. As shown in Figure~\ref{fig:butterfly_aspect_ratio}, {\ours} (blue curve) is discriminative to the change of aspect ratios while scores from the other ones trained with square resizing are not sensitive to the change. This shows that ARP resizing is important and \ours\ can effectively detect quality degradation due to resizing.

\vspace{+1.5mm}
\noindent\textbf{Effect of Full-size Input and the Multi-scale Input Composition.}
In Table \ref{tab:paq2piq-results} \ref{tab:koniq-results} \ref{tab:spaq-results} \ref{tab:ava-results}, we compare using only the full-size input (\fullours) and the multi-scale input (\ours). \fullours\ achieves promising results, showing the importance of preserving full-size input in IQA. The performance is further improved using multi-scale and the gain is larger on PaQ-2-PiQ and AVA because these two datasets have much more diverse resolutions than KonIQ-10k and SPAQ. This shows that multi-scale is important for effectively capturing quality information on real-world images with varying sizes.

We also vary the multi-scale composition and show in Table~\ref{tab:multi-scale-composition} that multi-scale consistently improves performance on top of single-scale models. The performance gain of multi-scale is more than a simple ensemble of individual scales because an average ensemble of individual scales actually under-performs using only the full-size image. Since \ours\ has full receptive field of the multi-scale input sequences, it can more effectively aggregate quality information across scales. 

\begin{table}[!tp]
\footnotesize
\begin{center}
\begin{tabular}{lrrr}\toprule
Multi-scale Composition &SRCC &PLCC \\\midrule
(224) &0.600 &0.667 \\
(384) &0.618 &0.695 \\
(512) &0.620 &0.691 \\
(384, 224) &0.620 &0.707 \\
(512, 384, 224) &0.629 &0.718 \\\midrule
(full) &0.640 &0.721 \\
(full, 224) &0.643 &0.726 \\
(full, 384) &0.642 &0.730 \\
(full, 384, 224) &\textbf{0.646} &\textbf{0.739} \\\midrule
Average ensemble of (full), (224), (384) &0.640 &0.710 \\
\bottomrule
\end{tabular}
\end{center}
\vspace{-2mm}
\caption{Comparison of multi-scale representation composition on PaQ-2-PiQ full-size test set. The multi-scale representation is composed of the resolutions shown in the brackets. Numbers in brackets indicate the longer side length $L$ for ARP resizing. "full" means full-size input image.  }\label{tab:multi-scale-composition}
\vspace{-3mm}
\end{table}

\begin{figure}[!tp]
\centering
\includegraphics[width=8cm]{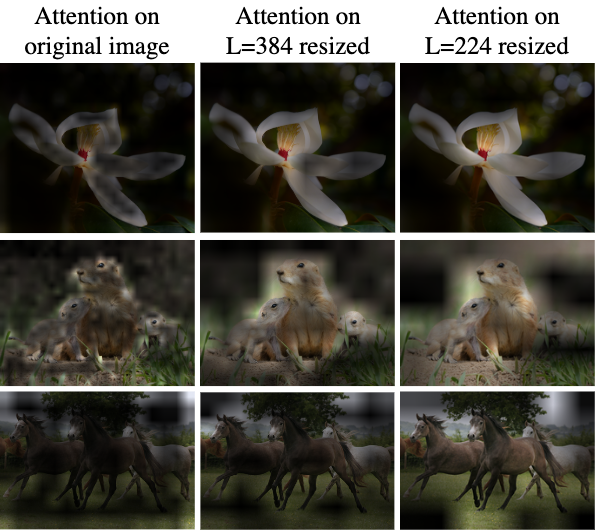}
\caption{Visualization of attention from the output tokens to the multi-scale representation (original resolution image and two ARP resized variants). Note that images here are resized to fit the grid, the model inputs are 3 different resolutions. The model is focusing on details in higher resolution image and on global area in lower resolution ones.}\vspace{-2mm}
\label{fig:attention_visualization} 
\end{figure}

To further verify that the model captures different information at different scales, we visualize the attention weights on each image in the multi-scale representation as Figure~\ref{fig:attention_visualization}. We observe that the model tends to focus on more detailed areas on full-size high-resolution images and on more global areas on the resized ones. This shows that the model learns to capture image quality at different granularities.

\begin{figure}[!tp]
\centering
\includegraphics[width=7cm]{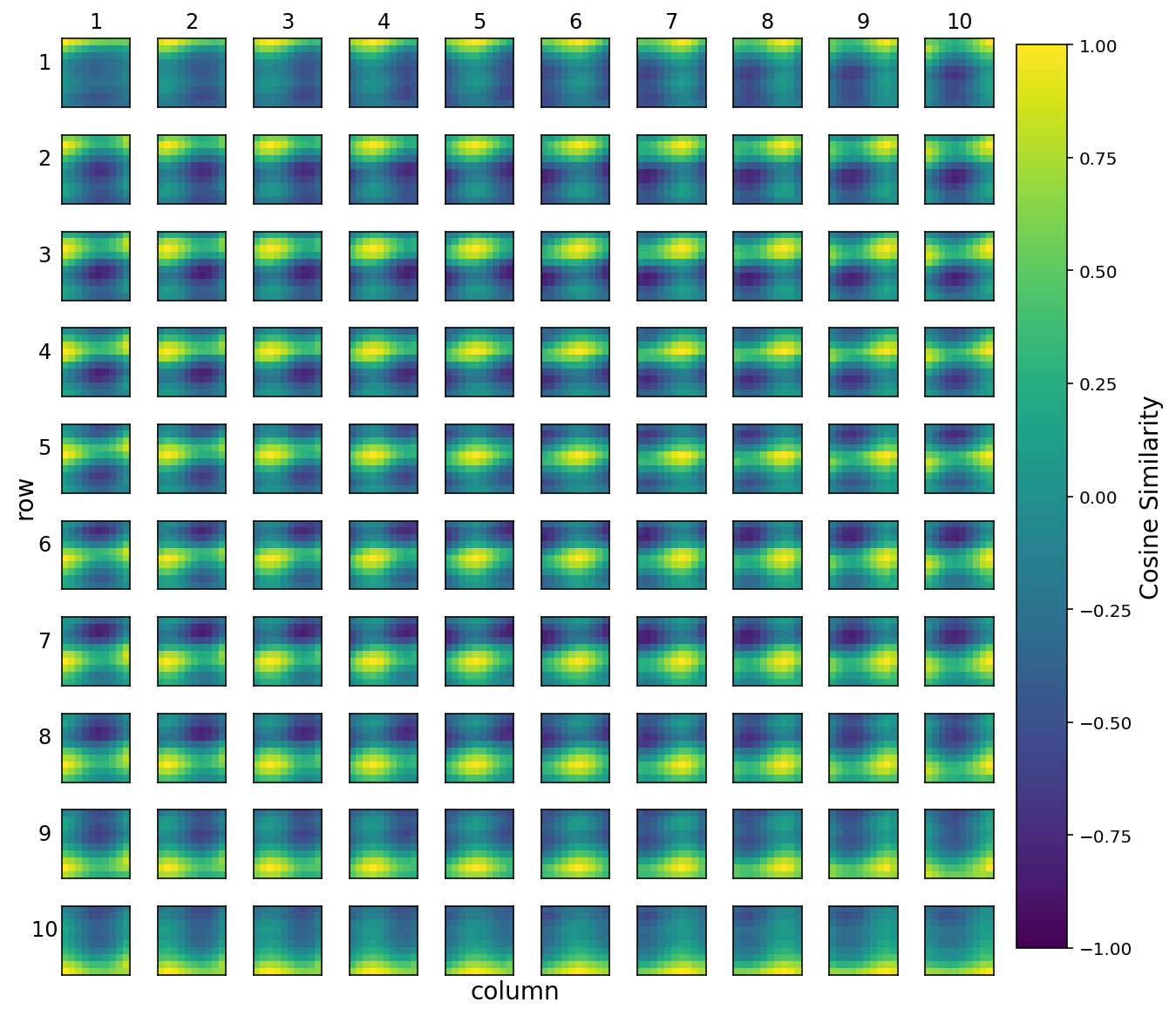}\vspace{-0.5mm}
\caption{Visualization of the grid of hash-based 2D spatial embedding with $G=10$. Each subplot ($i$, $j$) is of size $G\times G$, showing the cosine similarity between $T_{i,j}$ and every element in $T$. Visualizations for different $G$ are available in Appendix B.3.  }\vspace{-3mm}
\label{fig:hse_cosine_similarity_grid10} 
\end{figure}

\vspace{+1.5mm}
\noindent\textbf{Effectiveness of Proposed Hash-based Spatial Embedding (HSE) and Scale Embedding (SCE).} We run ablations on different ways to encode spatial information and scale information using positional embeddings. As shown in Table \ref{tab:spatial-emb}, there is a large gap between adding and not adding spatial embeddings. This aligns with the finding in \cite{dosovitskiy2020} that spatial embedding is crucial for injecting 2D image structure. To further verify the effectiveness of HSE, we try to add a fixed length spatial embedding as ViT \cite{dosovitskiy2020}. This is done by treating all input tokens as a fixed length sequence and assigning a learnable embedding for each position. The performance of this method is unsatisfactory compared to HSE because of two reasons: 1) the inputs are of different aspect ratios. So each patch in the sequence can come from a different location from the image. Fixed positional embedding fails to capture this change; 2) since each position is modeled independently, there is no cross-scale information, meaning that the model cannot locate spatially close patches from different scales in the multi-scale representation. Moreover, the method is inflexible because fixed length spatial embedding cannot be easily applied to the large images with more patches. On the contrary, HSE is meaningful under all conditions.

\begin{table}[!tp]
\footnotesize
\begin{center}
\begin{tabular}{cccc}\toprule
Spatial Embedding &SRCC &PLCC \\\midrule
w/o &0.704 &0.716 \\
Fixed-length (no HSE) &0.707 &0.722 \\
HSE &\textbf{0.726} &\textbf{0.738} \\
\bottomrule
\end{tabular}
\end{center}
\vspace{-2mm}
\caption{Ablation study results for spatial embeddings on AVA. For "Fixed length (not HSE)", we consider the input as a fixed-length sequence and assign a learnable embedding for each position.}\label{tab:spatial-emb}
\vspace{-1mm}
\end{table}

\begin{table}[!tp]
\footnotesize
\begin{center}
\begin{tabular}{cccc}\toprule
Scale Embedding &SRCC &PLCC \\\midrule
w/o &0.717 &0.729 \\
w/ &\textbf{0.726} &\textbf{0.738} \\
\bottomrule
\end{tabular}
\end{center}
\vspace{-2mm}
\caption{Ablation study results for scale embedding on AVA.}\label{tab:scale-emb}
\vspace{-2mm}
\end{table}

A visualization of the learned HSE cosine similarity is provided as Figure \ref{fig:hse_cosine_similarity_grid10}. As depicted, the HSE of spatially close locations are more similar (yellow color) and it corresponds well to the 2D structure. For example, the bottom HSEs are brightest at the bottom. This shows that HSE can effectively capture the 2D structure of the image.

In Table \ref{tab:scale-emb}, we show that adding SCE can further improve performance when compared with not adding SCE. This shows that SCE is helpful for the model to capture scale information independently of the spatial information.

\noindent\textbf{Choice of Patch Encoding Module.} We tried different designs for encoding the patch, including linear projection as \cite{dosovitskiy2020} and small numbers of convolutional layers. As shown in Table \ref{tab:patch-encoding}, using a simple convolution based patch encoding module can boost the performance. Adding more conv layers has diminishing returns and we find a 5-layer ResNet can provide satisfactory representation for the patch.


\begin{table}[!tp]
\footnotesize
\begin{center}
\begin{tabular}{lcccc}\toprule
&\# Params &SRCC &PLCC \\\midrule
Linear projection &22M &0.634 &0.714 \\
Simple Conv &23M &0.639 &0.726 \\
5-layer ResNet &27M &\textbf{0.646} &\textbf{0.739} \\
\bottomrule
\end{tabular}
\end{center}
\vspace{-2mm}
\caption{Comparison of different patch encoding modules on PaQ-2-PiQ full-size test set. For simple conv, we use the root of ResNet (a 7x7 conv followed by a 3x3 conv). For 5-layer ResNet, we stack a residual block on top of Simple Conv.}\label{tab:patch-encoding}
\vspace{-3mm}
\end{table}

\section{Conclusion}
\noindent We propose a multi-scale image quality Transformer (\ours), which can handle full-size image input with varying resolutions and aspect ratios. By transforming the input image to a multi-scale representation with both global and local views, the model is able to capture the image quality at different granularities. To encode positional information in the multi-scale representation, we propose a hash-based 2D spatial embedding and a scale embedding strategy. Although \ours\ is designed for IQA, it can be applied to other scenarios where task labels are sensitive to the image resolutions and aspect ratios. Moreover, \ours\ is compatible with any type of Transformers that accept input as a sequence of tokens. Experiments on the four large-scale IQA datasets show that \ours\ can consistently achieve state-of-the-art performance, demonstrating the effectiveness of the proposed method.

{\small
\setlength{\bibsep}{0pt}
\bibliography{egbib}
}

\clearpage
\appendix
\begin{center}
  \Large{\bf Appendix for ``MUSIQ: Multi-scale Image Quality Transformer'' }  
\end{center}

\section{Transformer Encoder}
\label{sec:transformer-encoder}
\subsection{Transformer Encoder Structure}
\begin{figure}[!htp]
\centering
\includegraphics[width=3cm]{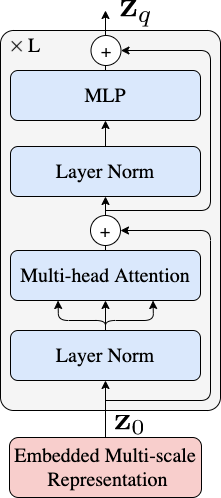}\vspace{+1mm}
\caption{Transformer encoder illustration. Graph inspired by \cite{NIPS2017_3f5ee243, dosovitskiy2020}. }\vspace{-2.5mm}
\label{fig:transformer-encoder} 
\end{figure}

\noindent We use the classic Transformer encoder \cite{NIPS2017_3f5ee243} in our experiments. As illustrated in Figure \ref{fig:transformer-encoder}, the Transformer block layer consists of multi-head self-attention (MSA), Layernorm (LN) and MLP layers. Residual connections are added in between the layers.

In \ours, the multi-scale patches are encoded as $\mathbf{x}_k^n$ where $k=0\cdots K$ is the scale index and $n$ is the patch index in the scale. $k=0$ represents the full-size image. We then add HSE and SCE to the patch embeddings, forming the multi-scale representation input. Similar to previous works \cite{devlin2018bert}, we prepend a learnable [class] token embedding to the sequence of embedded tokens ($\mathbf{x}_{class}$).

The Transformer encoder can be formulated as:
\begin{align}
    &\mathbf{E}_p = [\mathbf{x}_0^1; \cdots \mathbf{x}_0^l; \mathbf{x}_1^1; \cdots; \mathbf{x}_1^{m_1}; \cdots; \mathbf{x}_K^1; \cdots; \mathbf{x}_K^{m_K}] \\
    &\mathbf{z}_0 = [\mathbf{x}_{class}; \mathbf{E}_p + \mathbf{E}_{\text{HSE}}  + \mathbf{E}_{\text{SCE}}] \\
    &\mathbf{z}^{\prime}_q = \text{MSA}(\text{LN}(\mathbf{z}_{q-1})) + \mathbf{z}_{q-1},\ \ \ \ \ \ \ \ \ \  q =1\cdots L \\
    &\mathbf{z}_q = \text{MLP}(\text{LN}(\mathbf{z}^{\prime}_{q})) + \mathbf{z}^{\prime}_{q},\ \ \ \ \ \ \ \ \ \ \ \ \ \ \ \ \ \ q =1\cdots L \\
    &\mathbf{y} = \text{LN}(\mathbf{z}_L^0) &
\end{align}
$\mathbf{E}_p$ is the patch embedding. $\mathbf{E}_{\text{HSE}}$ and $\mathbf{E}_{\text{SCE}}$ are the spatial embedding and scale embedding respectively. $l$ is the number of patches from original resolution. $m_1 \cdots m_K$ are the number of patches from resized variants. $\mathbf{z}_0$ is the input to the Transformer encoder. $\mathbf{z}_q$ is the output of each Transformer layer and $L$ is the total number of Transformer layers.

\subsection{Multi-head Self-Attention (MSA)}
\begin{figure}[!htp]
\centering
\includegraphics[width=3cm]{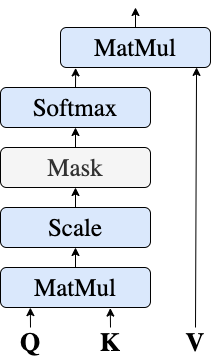}\vspace{+1mm}
\caption{Single head self-attention (SA) illustration. }\vspace{-2.5mm}
\label{fig:transformer-attention} 
\end{figure}

\noindent In this section we introduce the standard $\mathbf{QKV}$ self-attention (SA) \cite{NIPS2017_3f5ee243} (Figure \ref{fig:transformer-attention}) and its multi-head version (MSA). Suppose the input sequence is represented by $\mathbf{z} \in \mathbb{R} ^{N \times D}$, $\mathbf{Q, K, V}$ are its query, key, and value representations, respectively. They are generated by projecting the input sequence 
with a learnable matrix $\mathbf{U}_q, \mathbf{U}_k, \mathbf{U}_v \in \mathbb{R}^{D \times D_h}$, respectively. $D_h$ is the inner dimension for $\mathbf{Q, K, V}$. We then compute a weighted sum over $\mathbf{V}$ using attention weights $\mathbf{A} \in \mathbb{R} ^{N \times N}$ which are pairwise similarities between $\mathbf{Q}$ and $\mathbf{K}$.
\begin{align}
        \mathbf{Q} &= \mathbf{z}\mathbf{U}_q,\ \ \mathbf{K} = \mathbf{z}\mathbf{U}_k,\ \ \mathbf{V} = \mathbf{z}\mathbf{U}_v \\
        \mathbf{A} &= \text{softmax}(\mathbf{QK}^T / \sqrt{D_h}) \label{eq:softmax_a} \\
        \text{SA}(\mathbf{z}) &= \mathbf{AV}
\end{align}
MSA is an extension of SA where $s$ self-attention operations (heads) are conducted in parallel. The outputs from all heads are concatenated together and then projected to the final output with a learnable matrix $\mathbf{U}_m \in \mathbb{R}^{s\cdot D_h \times D}$. $D_h$ is typically set to $D / s$ to keep computation and number of parameters constant for each $s$.
\begin{equation}
    \text{MSA}(\mathbf{z}) = [\text{SA}_1(\mathbf{z}); \cdots; \text{SA}_s(\mathbf{z})]\mathbf{U}_m
\end{equation}

\subsection{Masked Self-Attention}
\noindent Masking is often used in self-attention \cite{NIPS2017_3f5ee243, devlin2018bert} to ignore padding elements or to restrict attention positions and prevent data leakage (\eg in causal or temporal predictions). In batch training, we use the input mask to indicate the effective input and to ignore padding tokens. As shown in Figure~\ref{fig:transformer-attention}, the mask is added on attention weights before the softmax. By setting the corresponding elements to $-\text{inf}$ before the softmax step in Equation \ref{eq:softmax_a}, the attention weights on invalid positions are close to zero.

The attention mask is constructed as  $\mathbf{M} \in \mathbb{R}^{N\times N}$ where 
\begin{equation}
    \mathbf{M}_{i, j} = 
    \begin{cases}
      0 & \text{if attention pos}_i \rightarrow \text{pos}_j \text{ valid} \\
      -\text{inf} & \text{if attention pos}_i \rightarrow \text{pos}_j \text{ invalid} \\
    \end{cases}    
\end{equation}
Then the masked self-attention weight matrix is calculated as

\begin{equation}
\mathbf{A}_m = \text{softmax}((\mathbf{QK}^T + \mathbf{M}) / \sqrt{D_h}).
\end{equation}

\subsection{Different Transformer Encoder Settings}

\noindent We use a lightweight parameters setting for Transformer encoder in the main experiments to make the model size comparable to ResNet-50. Here we also report the results from different Transformer encoder settings. The model variants are shown as Table~\ref{tab:transformer-variants}. The \ours-Small model is the one used in our main experiments in the paper. The performance of these variants on the AVA dataset is shown in Table~\ref{transformer-variants-ava-results}. Overall, these models have similar performance when pre-trained on ImageNet~\cite{imagenet}. Larger Transformer backbones might need more data to pre-train in order to get better performance. As shown in experiments from \cite{dosovitskiy2020}, larger Transformer backbones get better performance when pre-trained on ImageNet21k~\cite{deng2009imagenet} or JFT-300m~\cite{jftdataset}.

\begin{table}[!htp]
\footnotesize
\begin{center}
\setlength\tabcolsep{3.5pt}
\begin{tabular}{lcccccc}\toprule
& &Hidden size &MLP & & \\
Model &Layers & $D$ & size &Heads &Params\\\midrule
\ours-Small &14 &384 &1152 &6 &27M \\
\ours-Medium &8 &768 &2358 &8 &61M \\
\ours-Large &12 &768 &3072 &12 &98M \\
\bottomrule
\end{tabular}
\end{center}
\caption{\ours\ variants with different Transformer encoder settings.}\label{tab:transformer-variants}
\end{table}

\begin{table}[!htp]
\footnotesize
\begin{center}
\begin{tabular}{lccc}\toprule
Model &SRCC &PLCC \\\midrule
\ours-Small &0.916 &\textbf{0.928} \\
\ours-Medium &\textbf{0.918} &\textbf{0.928}  \\
\ours-Large &0.916 &0.927  \\
\bottomrule
\end{tabular}
\end{center}
\caption{Performance of different \ours\ variants on the KonIQ-10k dataset.}\label{transformer-variants-ava-results}
\end{table}


\vspace{+2mm}
\section{Additional Studies for HSE}
\subsection{Grid Size $G$ in HSE}
\label{sec:grid-size}

\noindent We run ablation studies for the grid size $G$ in the proposed hash-based 2D spatial embedding (HSE). Results are shown in Table \ref{tab:hse-grid-size}. Small $G$ may result in collision and therefore the model cannot distinguish spatially close patches. Large $G$ means the hashing is more sparse and therefore needs more diverse resolutions to train, otherwise some positions may not have enough data to learn good representations. One can potentially generate fixed $T$ for larger $G$ when detailed positions really matter (\eg using sinusoidal function, see Appendix \ref{sec:sinusoidal-hse}). With a learnable $T$, a good rule of thumb is to let grid size times the number of patches $P$ roughly equals the average resolution, \ie $G\times G \times P \times P = H \times W$. Since the average resolution across 4 datasets is around $450\times500$ and we use patch size 32, we use grid size around 10 to 15. Overall, we find different $G$ does not change the performance too much once it is large enough, showing that rough spatial encoding is sufficient for IQA tasks.

\begin{table}[!tp]
\footnotesize
\begin{center}
\begin{tabular}{lrrr}\toprule
Spatial Embedding &SRCC &LCC \\\midrule
HPE ($G=5$) &0.720 &0.733 \\
HPE ($G=8$) &0.723 &0.734 \\
HPE ($G=10$) &\textbf{0.726} &\textbf{0.738} \\
HPE ($G=12$) &0.722 & 0.736 \\
HPE ($G=15$) &0.724 & 0.735 \\
HPE ($G=20$) &0.722 & 0.734 \\
\bottomrule
\end{tabular}
\end{center}
\caption{Ablation study for different grid size $G$ in HSE on AVA dataset.}\label{tab:hse-grid-size}
\end{table}

\begin{table}[!tp]
\footnotesize
\begin{center}
\begin{tabular}{lccccc}\toprule
&\multicolumn{2}{c}{Learnable $T$} &\multicolumn{2}{c}{Fixed-Sin $T$} \\\midrule
G &SRCC &PLCC &SRCC &PLCC \\\midrule
10 &\textbf{0.726} &\textbf{0.738} &0.719 &\textbf{0.733} \\
15 &0.724 &0.735 &0.716 &0.730 \\
20 &0.722 &0.734 &\textbf{0.720} &\textbf{0.733} \\
\bottomrule
\end{tabular}
\end{center}
\caption{Comparison of sinusoidal HSE and learnable HSE matrix on 
AVA dataset.}\label{tab:sinusoidal-hse}
\end{table}

\subsection{Sinusoidal HSE v.s. Learnable HSE}
\label{sec:sinusoidal-hse}
\noindent Besides the learnable HSE matrix $T \in \mathbb{R} ^{G \times G \times D}$ introduced in the paper, another option is to generate a fixed positional encoding matrix $T$ using the sinusoidal function as \cite{NIPS2017_3f5ee243}. In Table~\ref{tab:sinusoidal-hse}, we show the performance comparison of using learnable $T$ or generated sinusoidal $T$ with different Grid size $G$.  Overall, the learnable $T$ gives slightly better performance than that of the fixed $T$.

\subsection{Visualization of HSE with Different $G$}
\label{sec:more-hse-visualiztion}

\begin{figure}[!tp]
\begin{subfigure}
\centering
\includegraphics[width=7cm]{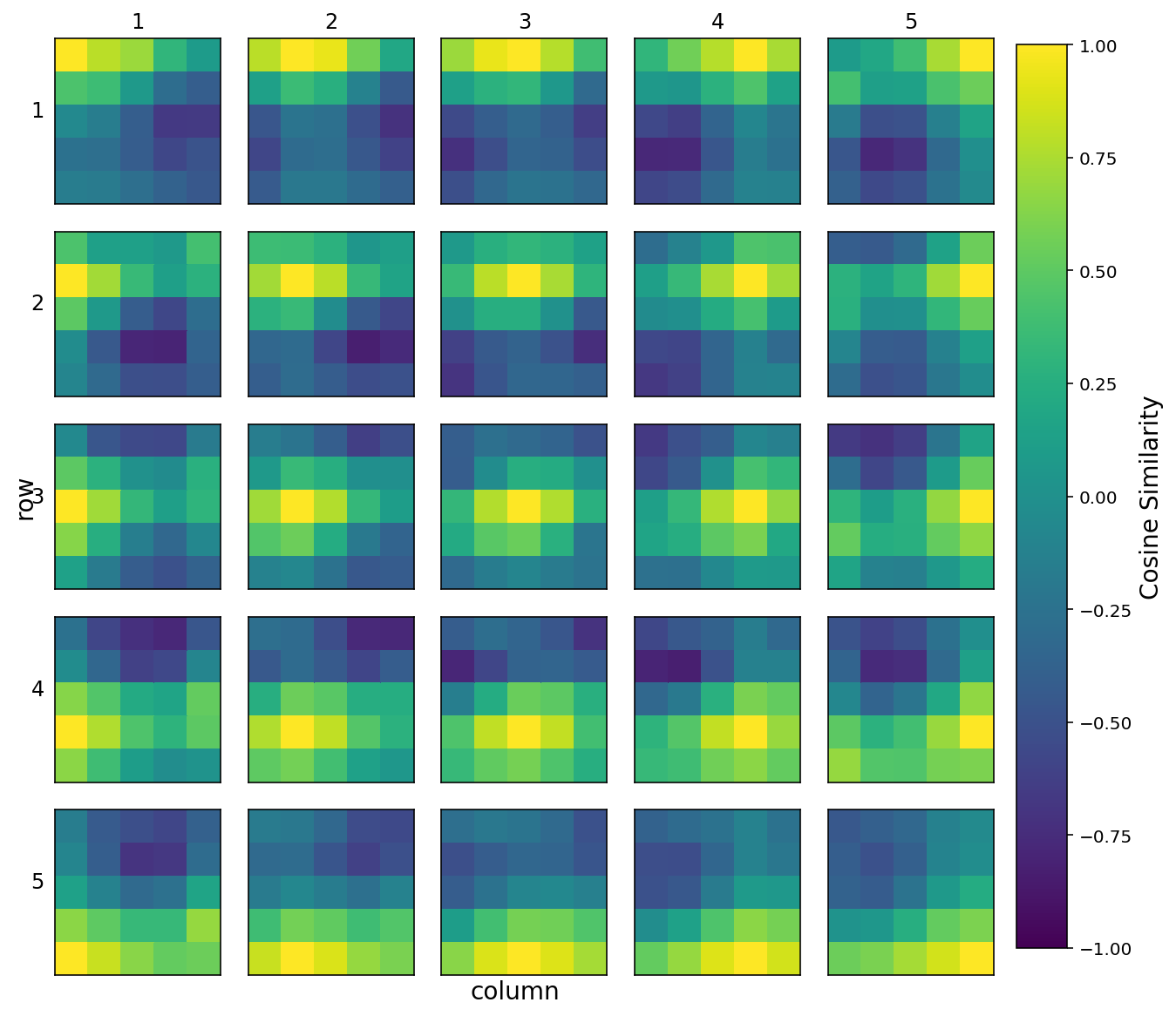}
\caption{Visualization of the grid of HSE with $G=5$.  }
\label{fig:hse_cosine_similarity_grid5} 
\end{subfigure}
\begin{subfigure}
\centering
\includegraphics[width=7cm]{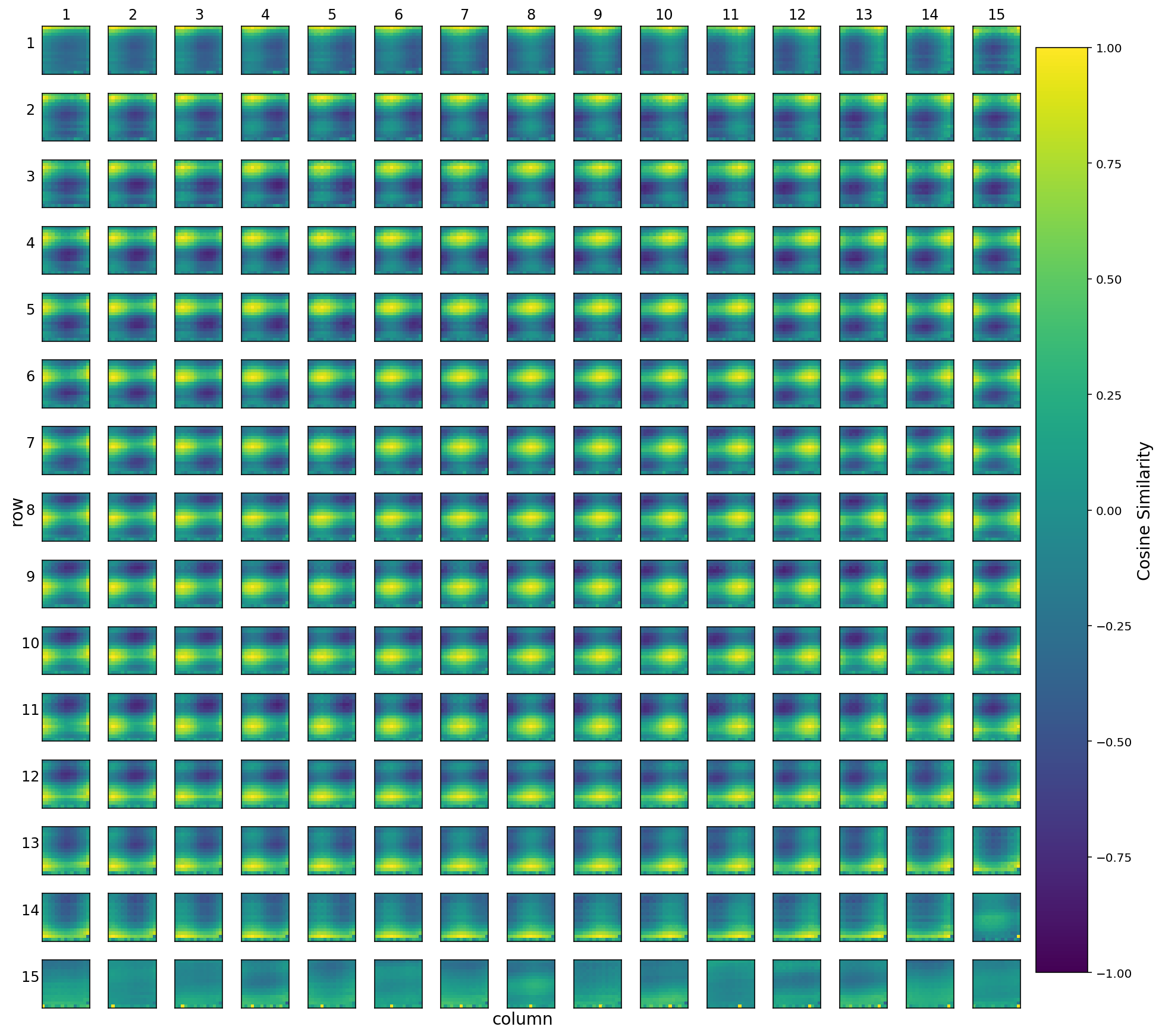}
\caption{Visualization of the grid of HSE with $G=15$. }
\label{fig:hse_cosine_similarity_grid15} 
\end{subfigure}
\end{figure}

\noindent Figures~\ref{fig:hse_cosine_similarity_grid5} and \ref{fig:hse_cosine_similarity_grid15} visualize the learned HSE with $G=5$ and $G=15$, respectively. Even with $G$ as small as 5, the similarity matrix corresponds well to the patch position in the image, showing that HSE captures patch position in the image.

\section{Effect of Patch Size}

We ran ablation on different patch size $P$, results are shown in Table \ref{tab:patch-size-ava}.  In our settings, we find patch size $P=32$ performs well across datasets.


\begin{table}[!tp]
\footnotesize
\begin{center}
\begin{tabular}{cccccc}\toprule
Patch Size &16 &32 &48 &64 \\\midrule
SRCC &0.715 &\textbf{0.726} &0.713 &0.705 \\
PLCC &0.729 &\textbf{0.738} &0.727 &0.719 \\
\bottomrule
\end{tabular}
\end{center}
\caption{Comparison of different patch size on AVA dataset.}\label{tab:patch-size-ava}
\vspace{-3mm}
\end{table}


\section{The Maximum Number of Patches ($l$) from Full-size Image }
\noindent We run ablation with different $l$ during training. As shown in the Table~\ref{tab:patch-length}, using large $l$ in the fine-tuning can improve the model performance. Since larger resolution images have more patches than low resolution ones,  when $l$ is too small, some larger images might be cutoff, thus the model performance will degrade.  

\begin{table}[!htp]
\footnotesize
\begin{center}
\begin{tabular}{lrrr}\toprule
$l$ &SRCC &LCC \\\midrule
128 &0.876 &0.895 \\
256 &0.906 &0.923 \\
512 &\textbf{0.916} &\textbf{0.928} \\
\bottomrule
\end{tabular}
\end{center}
\caption{Comparison of maximum \#patches $l$ from full-size image on KonIQ-10k dataset.}\label{tab:patch-length}
\end{table}

\section{KonIQ-10k More Results}
In our main experiment on KonIQ-10k, we followed BIQA~\cite{su2020blindly} and MetaIQA~\cite{zhu2020metaiqa} to report the average of 10 random 80/20 train-test splits to avoid the bias. On the other hand, methods like  KonCept512~\cite{hosu2020koniq} uses a fixed split instead of averaging. In Table~\ref{tab:compare-koncept512},  we report our results using the same fixed split. Images in KonIQ-10k are of the same resolution and CNN models like KonCept512 usually need a cherry-picked fixed size to work well.  Unlike CNN models that are constrained by fixed size, \ours\ does not need tuning the input size and generalizes well for diverse resolutions.

\begin{table}[!htp]
\footnotesize
\begin{center}
\begin{tabular}{lrrr}\toprule
method &SRCC &LCC \\\midrule
KonCept512~\cite{hosu2020koniq} &0.921 &\textbf{0.937} \\
\ours\ (Ours) &\textbf{0.924} &\textbf{0.937} \\
\bottomrule
\end{tabular}
\end{center}
\caption{Results on KonIQ-10k dataset using same fixed split as KonCept512~\cite{hosu2020koniq}.}\label{tab:compare-koncept512}
\end{table}

\section{SPAQ Full-size Results}
\noindent As mentioned in Section 4.1, we follow \cite{fang2020perceptual} to resize the raw images such that the shorter side is 512 for a fair comparison with the reference methods. Since our model can be applied directly on the images without resizing, we also report the performance on the SPAQ full-size test in Table~\ref{tab:full-size-train-spaq} when training on the SPAQ full-size train. The results only have very little difference.

\begin{table}[!htp]
\footnotesize
\begin{center}
\begin{tabular}{lccc}\toprule
&SRCC &PLCC \\\midrule
Full-size train and test &0.916 ($\pm0.001$) &0.919 ($\pm0.001$) \\
Resized train and test &0.917 ($\pm0.002$) &0.921 ($\pm0.002$) \\
\bottomrule
\end{tabular}
\end{center}
\caption{Comparison of \ours\ train and evaluate on full-size SPAQ dataset or the 512 shorter side resized SPAQ dataset.}\label{tab:full-size-train-spaq}
\end{table}

\section{Computation Complexity}
For the default \ours\ model, the number of parameters is around 27M. For a 224x224 image, its FLOPS is $8.86 \times 10^9$, which is at the same level as SOTA CNN-based models (23M parameters and $3.8 \times 10^9$ FLOPS for ResNet50). Training IQA takes 0.8 TPUv3-core-days on average. MST-IQA is compatible with the efficient Transformer backbones like Linformer and Performer, which greatly reduce the complexity of the original Transformer. We leave model speedup as the future work. 

\section{Multi-scale Attention Visualization}
\label{sec:more-attention-visualization}
\begin{figure*}[!htp]
\centering
\includegraphics[width=17cm]{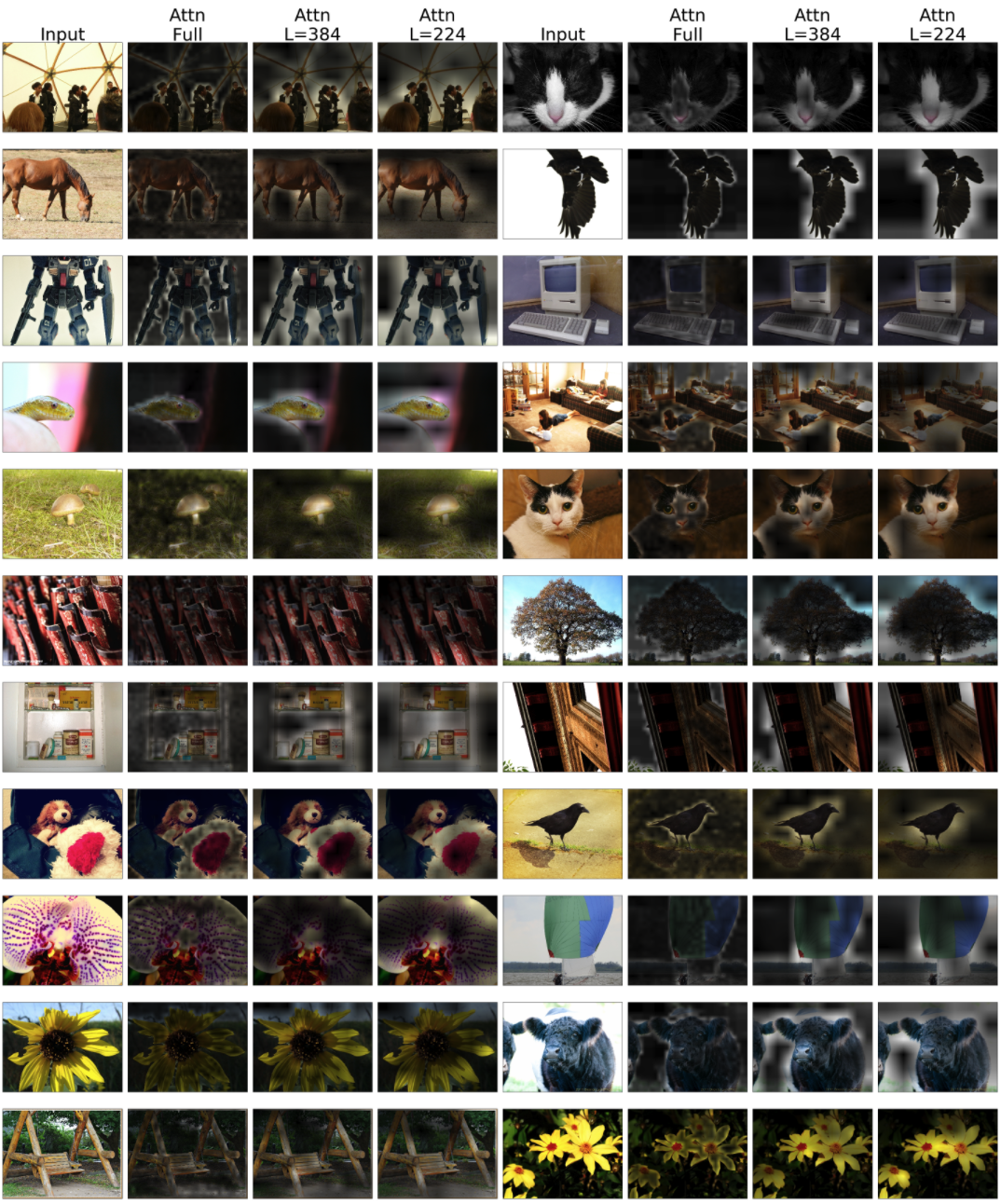}\vspace{-0.5mm}
\caption{Visualizations of attention from the output tokens to the multi-scale representation. ``Input" column shows the input image. ``Attn Full" shows the attention on the full-size image. ``Attn L=384" and ``Attn L=224" show the attention on the ARP resized images. Note that images here are resized to fit the grid, the model inputs are 3 different resolutions.}\vspace{-2.5mm}
\label{fig:more-attention-visualization} 
\end{figure*}

\noindent To understand how \ours\ uses self-attention to integrate information across different scales, we visualize the average attention weights from the output tokens to each image in the multi-scale representation as  Figure~\ref{fig:more-attention-visualization}. We follow \cite{dosovitskiy2020} for the attention map computation. In short, the attention weights are averaged across all heads and then recursively multiplied, accounting for the mixing of attention across tokens through all layers.

\clearpage
\end{document}